\documentclass[english]{IEEEtran}
\usepackage[T1]{fontenc}
\usepackage[latin9]{inputenc}
\usepackage{authblk}
\usepackage{color}
\usepackage{array}
\usepackage{multirow}
\usepackage{amsmath}
\usepackage{amstext}
\usepackage{graphicx}
\usepackage[pdftex, colorlinks=true, hyperfootnotes=true, hyperindex=true,
            plainpages=false, pagebackref=false, pdfpagelabels=true, pdfstartview=FitH,
            linkcolor=blue, citecolor=blue, urlcolor=blue,
            bookmarks, bookmarksopen, bookmarksdepth=3]{hyperref}
\usepackage[capitalise,nameinlink]{cleveref}

\makeatletter

\providecommand{\tabularnewline}{\\}

\usepackage{algolyx}
\usepackage{eso-pic}
       
\@ifundefined{showcaptionsetup}{}{%
 \PassOptionsToPackage{caption=false}{subfig}}
\usepackage{subfig}
\usepackage{varwidth}
\makeatother
\usepackage{babel}

\begin{document}
\AddToShipoutPictureBG*{%
  \AtPageLowerLeft{%
    \setlength\unitlength{1in}%
    \hspace*{\dimexpr0.5\paperwidth\relax}
    \makebox(0,0.75)[c]{\small Paper communicated to 42nd IEEE Annual Conference of Engineering in Medicine and Biology Society, Montreal, Canada}%
}}
\title{Understanding patient complaint characteristics using contextual clinical BERT embeddings}
\author{Budhaditya Saha, Sanal Lisboa, Shameek Ghosh }
\affil{Medius Health, Sydney, Australia\\
\textit {\{aditya.saha, sanal.lisboa, shameek.ghosh\}@mediushealth.org}}
\maketitle

\global\long
\global\long\def\mF{\mathcal{F}}%

\global\long
\global\long\def\mA{\mathcal{A}}%

\global\long
\global\long\def\mH{\mathcal{H}}%

\global\long
\global\long\def\mX{\mathcal{X}}%

\global\long
\global\long\def\dist{d}%

\global\long
\global\long\def\HX{\entro\left(X\right)}%
 \global\long
\global\long\def\entropyX{\HX}%

\global\long
\global\long\def\HY{\entro\left(Y\right)}%
 \global\long
\global\long\def\entropyY{\HY}%

\global\long
\global\long\def\HXY{\entro\left(X,Y\right)}%
 \global\long
\global\long\def\entropyXY{\HXY}%

\global\long
\global\long\def\mutualXY{\mutual\left(X;Y\right)}%
 \global\long
\global\long\def\mutinfoXY{W\mutualXY}%

\global\long
\global\long\def\given{\mid}%

\global\long
\global\long\def\gv{\given}%

\global\long
\global\long\def\goto{\rightarrow}%

\global\long
\global\long\def\asgoto{\stackrel{a.s.}{\longrightarrow}}%

\global\long
\global\long\def\pgoto{\stackrel{p}{\longrightarrow}}%

\global\long
\global\long\def\dgoto{\stackrel{d}{\longrightarrow}}%

\global\long
\global\long\def\ll{\mathit{l}}%

\global\long
\global\long\def\logll{\mathcal{L}}%

\global\long
\global\long\def\bzero{\vt0}%

\global\long
\global\long\def\bone{\mathbf{1}}%

\global\long
\global\long\def\bff{\vt f}%

\global\long
\global\long\def\bx{\mathbf{x}}%

\global\long
\global\long\def\bX{\mathbf{X}}%

\global\long
\global\long\def\bW{\mathbf{W}}%

\global\long
\global\long\def\bM{M}%

\global\long
\global\long\def\bD{\mathbf{D}}%

\global\long
\global\long\def\bLambda{\boldsymbol{\Lambda}}%

\global\long
\global\long\def\bE{\mathbf{E}}%

\global\long
\global\long\def\bP{\mathbf{P}}%

\global\long
\global\long\def\bH{\mathbf{H}}%

\global\long
\global\long\def\bA{\mathbf{A}}%

\global\long
\global\long\def\bB{\mathbf{B}}%

\global\long
\global\long\def\bS{\mathbf{S}}%

\global\long
\global\long\def\bs{s}%

\global\long
\global\long\def\bG{\mathbf{G}}%

\global\long
\global\long\def\bR{\mathbf{R}}%

\global\long
\global\long\def\bL{\mathbf{L}}%

\global\long
\global\long\def\bI{\mathbf{I}}%

\global\long
\global\long\def\tbx{\tilde{\bx}}%

\global\long
\global\long\def\by{\mathbf{y}}%

\global\long
\global\long\def\bc{\mathbf{c}}%

\global\long
\global\long\def\br{\mathbf{r}}%

\global\long
\global\long\def\be{\mathbf{e}}%

\global\long
\global\long\def\bb{\mathbf{b}}%

\global\long
\global\long\def\bbm{\mathbf{m}}%

\global\long
\global\long\def\bh{\mathbf{h}}%

\global\long
\global\long\def\bn{\mathbf{n}}%

\global\long
\global\long\def\bg{\mathbf{g}}%

\global\long
\global\long\def\bd{\mathbf{d}}%

\global\long
\global\long\def\bw{\mathbf{w}}%

\global\long
\global\long\def\bY{\mathbf{Y}}%

\global\long
\global\long\def\bI{\mathbf{I}}%

\global\long
\global\long\def\bz{\mathbf{z}}%

\global\long
\global\long\def\bZ{\mathbf{Z}}%

\global\long
\global\long\def\bu{\mathbf{u}}%

\global\long
\global\long\def\bU{\mathbf{U}}%

\global\long
\global\long\def\bK{\mathbf{K}}%

\global\long
\global\long\def\bC{\mathbf{C}}%

\global\long
\global\long\def\bv{\mathbf{v}}%

\global\long
\global\long\def\bV{\mathbf{V}}%

\global\long
\global\long\def\balpha{\gvt\alpha}%

\global\long
\global\long\def\bbeta{\gvt\beta}%

\global\long
\global\long\def\bmu{\gvt\mu}%

\global\long
\global\long\def\btheta{\boldsymbol{\theta}}%

\global\long
\global\long\def\bOmega{\boldsymbol{\Omega}}%

\global\long
\global\long\def\blambda{\boldsymbol{\lambda}}%

\global\long
\global\long\def\realset{\mathbb{R}}%

\global\long
\global\long\def\realn{\real^{n}}%

\global\long
\global\long\def\natset{\integerset}%

\global\long
\global\long\def\interger{\integerset}%

\global\long
\global\long\def\integerset{\mathbb{Z}}%

\global\long
\global\long\def\natn{\natset^{n}}%

\global\long
\global\long\def\rational{\mathbb{Q}}%

\global\long
\global\long\def\realPlusn{\mathbb{R_{+}^{n}}}%

\global\long
\global\long\def\comp{\complexset}%
 \global\long
\global\long\def\complexset{\mathbb{C}}%

\global\long
\global\long\def\and{\cap}%

\global\long
\global\long\def\compn{\comp^{n}}%

\global\long
\global\long\def\comb#1#2{\left({#1\atop #2}\right) }%

\global\long
\global\long\def\argmin#1{\underset{#1}{\text{argmin}}}%

\global\long
\global\long\def\diag{\text{diag}}%

\begin{abstract}
In clinical conversational applications, extracted entities tend to capture the main subject of a  patient's complaint, namely symptoms or diseases. However, they mostly fail to recognize the characterizations of a complaint such as the time, the onset, and the severity. For example, if the input is ``I have a headache and it is extreme'', state-of-the-art models only recognize the main symptom entity - \emph{headache}, but ignore the severity factor of \emph{extreme}, that characterises \emph{headache}. 
In this paper, we design a two-stage approach to
detect the characterizations of entities like symptoms presented by general users in contexts where they would describe their symptoms to a clinician. We use Word2Vec and BERT to
encode clinical text given by the patients. We transform the output and re-frame the task as a multi-label classification problem. Finally, we combine the processed encodings with the Linear Discriminant Analysis
(LDA) algorithm to classify the characterizations of the main entity.
Experimental results demonstrate that our method achieves 40-50\% improvement
in the accuracy over the state-of-the-art models.
\end{abstract}

\section{Introduction\label{sec:intro}}

Clinical Named Entity Recognition systems based on neural networks \cite{liu2017entity} \cite{Neumann_2019} are trained to detect entities in text. In the clinical domain, there are different types of inter-related entities. Existing systems lack the ability to detect these relations because these systems are not trained to understand the context in a text.
 For example, in the text ''I have severe headache and nausea'', the parent entities are \emph{headache} and \emph{nausea}. The child of \emph{headache} is \emph{severe}. Existing systems may detect the three entities but they are unable to predict if they are related.
 
A relationship prediction mechanism is required to link parent and child entities in a text.
Such techniques are useful in applications like clinical conversational chat platforms \cite{ghosh2018quro}, which predict disease differentials based on the symptoms entered. Here, the quality of predicted disease differentials depends on the accuracy of identified clinical information ( and their characteristics) in the text. The input text may contain two main
components (a) clinical entities or \emph{parent}
entities and (b) the characterization of the clinical entities or \emph{children} entities. 
\cref{tab:Examples_TOS}
shows example of duration, severity, onset and frequency onset characterizations respectively.

The clinical named entity recognition model has been researched extensively
\cite{liu2017entity,Neumann_2019,zhang2013unsupervised}. The state-of-the-art
clinical entity recognition methods \cite{aronson2006metamap,bhatia2019comprehend}
mostly recognizes the parent entity in the text. The two most popular
clinical entity recognition tools are METAMAP \cite{aronson2006metamap}
and Amazon Medical Comprehend \cite{bhatia2019comprehend}. The METAMAP framework recognizes a medical concept, whereas the Amazon Medical Comprehend service predicts the named entities in clinical texts. While these tools can separately detect the parent entity and child entities, they are not able to predict the relationship or context. In the example discussed earlier, the Amazon medical comprehend predicts
the \emph{headache} and \emph{nausea} as a \emph{parent}
entities and \emph{severe} as a characterization. Similar outcomes
can also be found for the METAMAP. But they fail to
recognize that the \emph{severe} is related to \emph{headache}. The main reason behind this failure
is that the data modeling method ignores the contextual information
that denote the relation between the neighboring words. 

In this paper we build a solution to recognize the time, onset and severity
characterization of a \emph{parent}
entity in user input. In a clinical chatbot, the \emph{parent}
entities are mostly symptoms 
or diseases 
To achieve this,
we seek to capture the contextual information in an input text and
convert this information into a vector space representation. These
vector space model will effectively map contextually related texts
close to each other and unrelated texts far away from each other.
For example, sentences like \emph{I have a pain in the
head for hours} and \emph{I have got
a headache since morning} have similar
contextual information for a target entity \emph{headache} ,
hence, they will be placed close to each into the vector space. Similarly,
examples like \emph{I am having continuous headache} and \emph{I get headache infrequently}
will be mapped far away from each other, as confirmed by our experiments. 
\begin{figure*}
\centering{}\includegraphics[width=0.7\linewidth,height=0.30\linewidth]{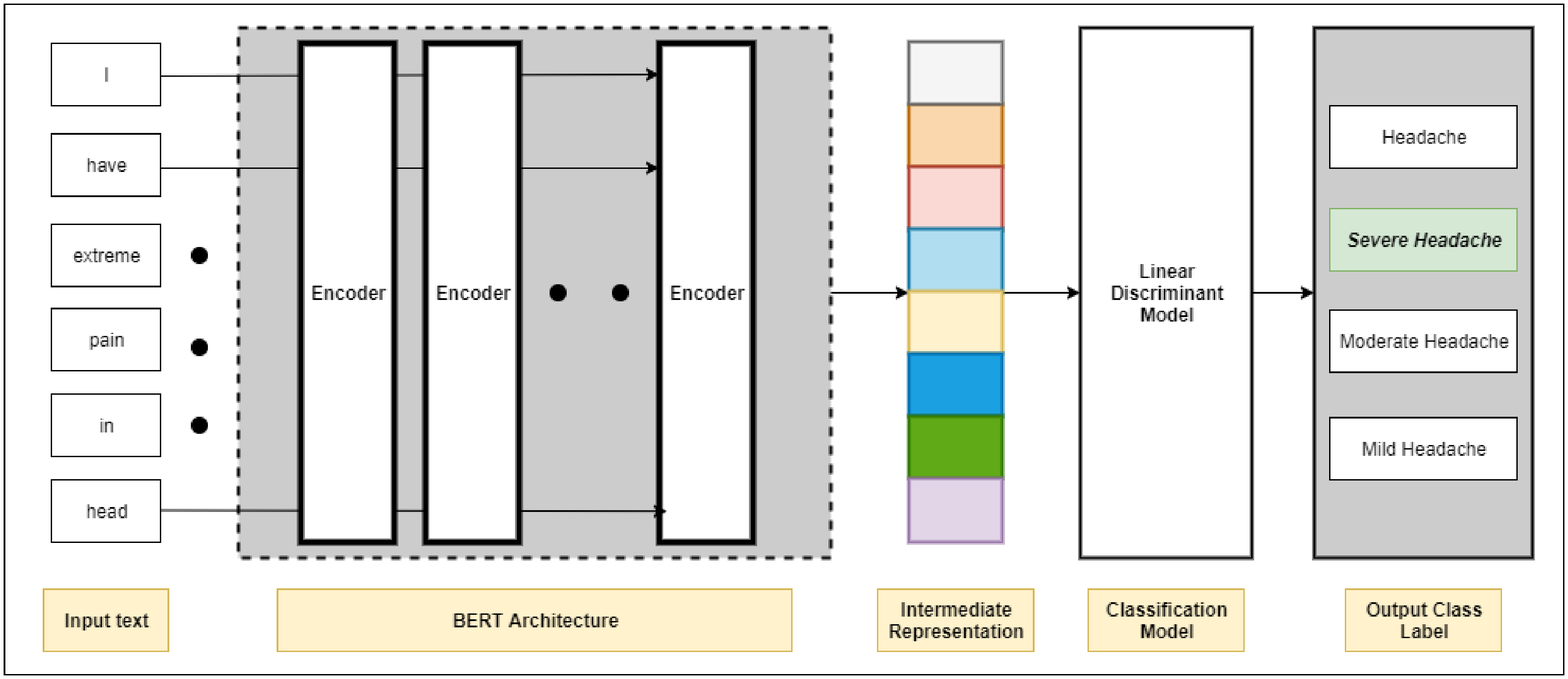}\caption{A schematic diagram of the proposed time-severity-onset recognition
model \label{fig:A-Schematic-diagram}}
\end{figure*}

Here, we propose a framework to understand the language of
the clinical text and predict the time, onset and severity characterization
of a \emph{parent} entity. For language
understanding, we use 
state-of-the-art, deep learning models designed for natural language processing tasks. 
In
our model, we have fine-tuned these models for the clinical text. These models take text as an input and output a continuous
vector representation. This outcome is processed and the downstream task is framed
as a multi-label classification problem. Finally, the intermediate representation
is fed to a classifier. We have applied the model on
the dataset curated from potential users of chatbot Quro \cite{ghosh2018quro}.
We compare the outcome of this model with other popular state-of-the-art
\cite{aronson2006metamap,bhatia2019comprehend} architectures. The
performance evaluation shows that our model archives up to 50\% higher accuracy, 40\% higher precession, 40\% higher recall, and 30\% higher F1 score compared to state-of-the-art.
\begin{table}[h]
\begin{centering}
\subfloat[Duration (time) \label{tab:duration}]{\centering{}%
\begin{tabular}{|>{\centering}m{3.5cm}|>{\centering}m{1.2cm}|>{\centering}p{1.3cm}|>{\centering}p{1cm}|}
\hline 
\textbf{User Input in a chatbot} & \textbf{Parent Entity} & \textbf{Children type} & \textbf{Children Class}\tabularnewline
\hline 
I have a \textbf{\textcolor{blue}{\emph{headache}}} for the last
\textbf{\textcolor{red}{\emph{2 months}}}\textbf{\textcolor{black}{\emph{.}}} & \multirow{4}{1.2cm}{Symptom ( Headache)} & \multirow{4}{1cm}{\textbf{Duration (time)}} & Months\tabularnewline
\cline{1-1} \cline{4-4} 
She is having a \textbf{\textcolor{blue}{\emph{headache}}} since last
\textbf{\textcolor{red}{\emph{five days}}}. &  &  & Days\tabularnewline
\cline{1-1} \cline{4-4} 
\textbf{\textcolor{blue}{\emph{headache}}} lasted for \textbf{\textcolor{red}{\emph{several
hours}}}. &  &  & Hours\tabularnewline
\cline{1-1} \cline{4-4} 
I'm having \textbf{\textcolor{blue}{\emph{headache}}} from \textbf{\textcolor{red}{\emph{few
minutes}}}\textbf{\textcolor{black}{\emph{.}}} &  &  & Minutes\tabularnewline
\hline 
\end{tabular}}
\par\end{centering}
\begin{centering}
\subfloat[Severity \label{tab:severity}]{\centering{}%
\begin{tabular}{|>{\centering}m{3.5cm}|>{\centering}m{1.2cm}|>{\centering}p{1.3cm}|>{\centering}p{1cm}|}
\hline 
\textbf{User Input in a chatbot} & \textbf{Parent Entity} & \textbf{Children type} & \textbf{Children Class}\tabularnewline
\hline 
\textbf{\textcolor{blue}{\emph{Pain}}} is \textbf{\textcolor{red}{\emph{extreme}}}
\textbf{\textcolor{blue}{\emph{in}}} my \textbf{\textcolor{blue}{\emph{head}}} & \multirow{3}{1.2cm}{Symptom ( Headache)} & \multirow{3}{1cm}{\textbf{Severity}} & Severe\tabularnewline
\cline{1-1} \cline{4-4} 
I am having a \textbf{\textcolor{red}{\emph{moderate}}} \textbf{\textcolor{blue}{\emph{headache}}} &  &  & Moderate\tabularnewline
\cline{1-1} \cline{4-4} 
I am having a \textbf{\textcolor{red}{\emph{slight}}}\textcolor{green}{{}
}\textbf{\textcolor{blue}{\emph{pain in head}}} &  &  & Mild\tabularnewline
\hline 
\end{tabular}}
\par\end{centering}
\begin{centering}
\subfloat[Onset \label{tab:onset}]{\centering{}%
\begin{tabular}{|>{\centering}p{3.5cm}|>{\centering}p{1.2cm}|>{\centering}p{1.3cm}|>{\centering}p{1cm}|}
\hline 
\textbf{User Input in a chatbot} & \textbf{Parent Entity} & \textbf{Children type} & \textbf{Children Class}\tabularnewline
\hline 
my \textbf{\textcolor{blue}{\emph{headache}}} starts \textbf{\textcolor{red}{\emph{abruptly}}} & \multirow{2}{1.3cm}{Symptom (Headache)} & \multirow{2}{1cm}{\textbf{Onset}} & Sudden\tabularnewline
\cline{1-1} \cline{4-4} 
\textbf{\textcolor{red}{\emph{gradual}}} \textbf{\textcolor{blue}{\emph{pain
in}}} my \textbf{\textcolor{blue}{\emph{head}}} &  &  & Gradual\tabularnewline
\hline 
\end{tabular}}
\par\end{centering}
\begin{centering}
\subfloat[Frequency (time) \label{tab:frequency}]{\centering{}%
\begin{tabular}{|>{\centering}p{3.5cm}|>{\centering}p{1.4cm}|>{\centering}p{1.3cm}|>{\centering}p{1cm}|}
\hline 
\textbf{User Input in a chatbot} & \textbf{Parent Entity} & \textbf{Children type} & \textbf{Children Class}\tabularnewline
\hline 
I am having a \textbf{\emph{\textcolor{red}{constant}}} \textbf{\emph{\textcolor{blue}{pain in head}}} & \multirow{2}{1.5cm}{Symptom (e.g. Headache)} & \multirow{2}{1cm}{\textbf{frequency (time)}} & Continuous\tabularnewline
\cline{1-1} \cline{4-4} 
I usually get \textbf{\emph{\textcolor{blue}{pain in head}}} \textbf{\emph{\textcolor{red}{occasionally}}} &  &  & ON - OFF\tabularnewline
\hline 
\end{tabular}}
\par\end{centering}
\centering{}\caption{Examples of duration, severity, onset, and frequency characterizations
respectively. The color represents the type of the entities (\emph{blue}
for parent and \emph{red} for children). \label{tab:Examples_TOS}}
\end{table}

\section{Clinical Dialogue Dataset }

The clinical conversation data is curated from text inputs of potential users of Quro \cite{ghosh2018quro}. The Quro bot is an AI-driven
clinical conversational platform that orchestrates the patient throughout
the primary health care journey. We have recorded around 2000 instances of unlabeled text which contains \emph{parent} symptoms with one or more \emph{time-severity-onset} factors for each parent. 
We use a semi-supervised technique to label these unlabeled text instances. The semi-supervised technique detects keywords in the text and labels them. These labels are then reviewed by clinicians to reduce the noise. \cref{tab:Examples_TOS} shows recorded data with annotations.

The time factor has two components ie. duration and frequency. There are 4 characteristics (hours, days, weeks, months) of duration component and 2 characteristics (on-off, continuous) of frequency component as shown in  \cref{tab:duration} and \Cref{tab:frequency}. If user text is \emph{I have been having regular back pain since the last 3 days}, the word \emph{regular} in text signifies  \emph{continuous} characteristic of frequency component; \emph{last 3 days} signifies the \emph{days} characteristic of duration component.

The severity factor has 3 characteristics ie. severe, mild and moderate as shown in \cref{tab:severity}. For example if the user text is \emph{I have extreme headache} the word \emph{extreme} in text signifies \emph{severe} characteristic.

The onset factor has 2 characteristics ie. sudden and gradual as shown in \cref{tab:onset}. For example if the user text is \emph{my headache starts abruptly} the word \emph{abruptly} in text signifies \emph{sudden} characteristic.


\section{Proposed Approach \label{sec:framework}}

The proposed framework has two main components: (a) A text encoder, and (b) a classification model. In our architecture, a tokenized sentence
is given as an input to the encoder which maps the
sentence to a continuous vector representation. The dimensions of the vector representation are reduced because most encoders produce high dimensional vector representations. Then,
the classification model maps the vector representation
to a child class. \cref{fig:A-Schematic-diagram} depicts a schematic
diagram of the proposed model which uses BERT as the encoder and LDA as the classifier. Before the classifier is applied the task is framed as a multi label classification problem. In the following sections, we will describe the
two components in detail.

\subsection{Text Encoding}

Word embeddings are vectors that represent words in a text in the semantic space. Similarity between words can be found using a distance measure such as cosine similarity.
The Word2Vec is a deep learning model which attempts to create high quality word embeddings \cite{mikolov2013distributed}. These word embeddings capture context form the training corpus and as such embeddings from general text cannot be used in highly specific context.

Bidirectional Encoder Representations from Transformer (BERT) \cite{devlin2018bert}
is a viable solution to the context problem. BERT model reads a text input
sequentially from left-to-right and right-to-left to learn the contextual
relationship amongst the words and embed this learning into a low
dimensional continuous vector space. This characteristic allows the
model to learn the context of a word based on all of its surroundings
(left and right of the word).

In our framework, we can use Word2Vec to encode words in sentence and combine the word vectors. Alternatively, we can use 
BERT encoder model to get a continuous vector representation of the entire input
text. These models can be pre-trained on large corpus of text and used in a different context. However the accuracy would depend on the relevance and quality of the corpus.

\subsection{Sentence Classification }

We use LDA for sentence classification. Consider, $C$ number of children class denoted by $\mathcal{{L}}$,
where $\mathcal{{L}}$ = ($\ell_{1},\ell_{2},\ldots,\ell_{m}$), the
LDA \cite{zhang2007discriminant} model maps the $D$ dimensional vector representation $\bh\in\mathcal{R}^{D}$ to a class label in $\mathcal{L}.$
Here, the training data can be expressed as $(\bh_{i},\ell_{i})$
where $i\in{1,\ldots N}$, $N$ is the number of training samples. The
number of vectors in class $\ell_{i}$ is denoted by $n_{i}$, thus
$N=\sum n_{i}.$ The LDA tries to find an optimal hyperplane such
that the separability between two classes is maximized. The hyperplane
is computed by minimizing the within class distance and maximizing
the between class distance simultaneously. The within class ($\bH_{w}$)
and between class ($\bH_{b}$) scatter matrices are defined as 

\begin{eqnarray*}
\bH_{b}=\sum_{i=1}^{C}(\bbm_{i}-\bbm)(\bbm_{i}-\bbm)^{T}\\
\bH_{w}=\sum_{i=1}^{C}\sum_{\bh\in\ell_{i}}(\bh-\bbm_{i})(\bh-\bbm_{i})^{T}
\end{eqnarray*}
where $\bbm_{i}$ denotes the class mean of $i^{th}$ class and $\bbm$
is the global mean of samples $\{\bh_{i}\}_{i=1}^{N}$. The LDA model
learns the hyperplane by optimizing the fisher criterion as 

\begin{eqnarray}
J(\bW)=\max_{\bW}\frac{\bW^{T}\bH_{b}\bW}{\bW^{T}\bH_{w}\bW}
\label{eq:lda}
\end{eqnarray}
where $\bW$ is a parameter of the hyperplane. The equation (1)
can also be modified as $\bH_{b}\bW=\lambda\bH_{w}\bW$ which turns
to a generalized eigenvalue problem with eigenvectors $\bW$ and eigenvalue
$\lambda$. The optimal hyperplane is spanned by the eigenvectors
in $\bW$.

\begin{figure*}[t]
    \centering
    \subfloat[Duration (Time)]{\includegraphics[width=0.47\textwidth]{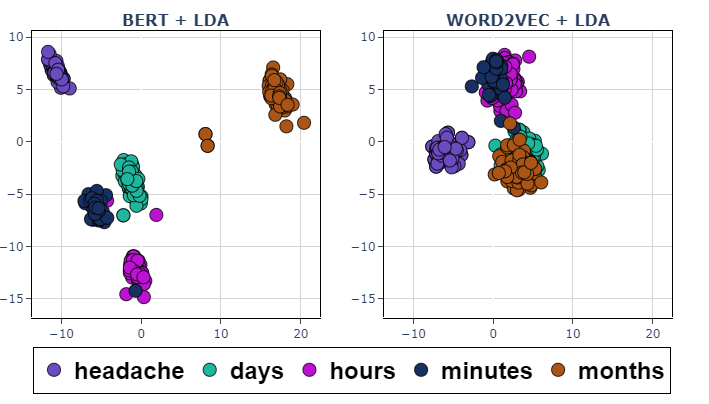}}
    \subfloat[Frequency (time)]{\includegraphics[width=0.47\textwidth]{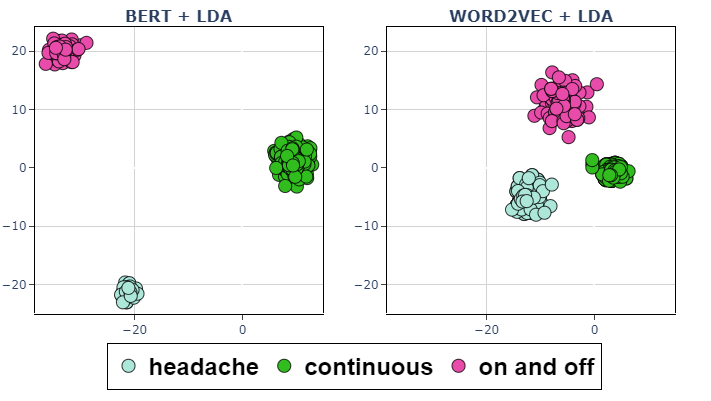}}
    \newline
    \centering
    \subfloat[Severity]{\includegraphics[width=0.47\textwidth]{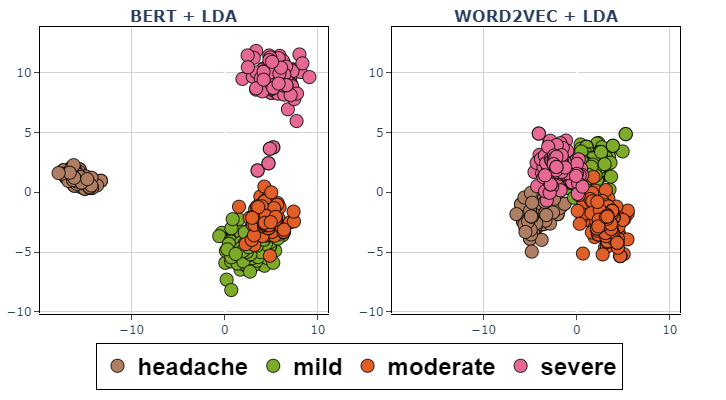}}
    \subfloat[Onset]{\includegraphics[width=0.47\textwidth]{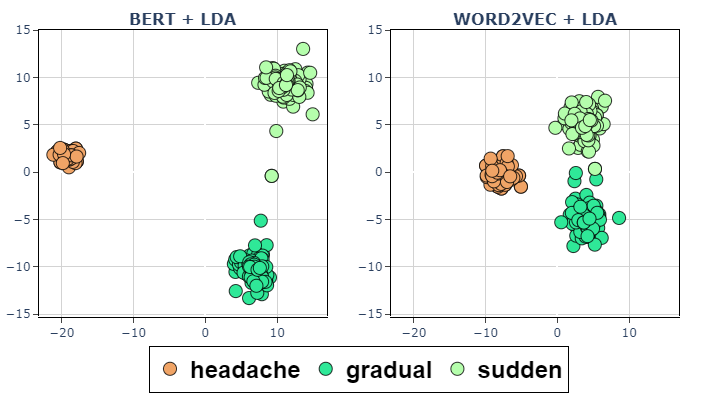}}
    \caption{Clusters formed by the BERT+LDA and Word2Vec+LDA
models.}
\label{fig:clusters}
\end{figure*}

\section{Experiments \label{sec:Experiments}}

\begin{table}
\begin{centering}
\begin{tabular}{|>{\centering}p{2cm}|c|c|c|c|}
\hline 
Model & Accuracy & Precession & Recall & F1-score\tabularnewline
\hline 
BERT+ LDA & \textbf{0.642} & \textbf{0.9245} & \textbf{0.8376} & \textbf{0.8789}\tabularnewline
\hline 
Amazon Medical  & 0.1714 & 0.5982 & 0.3418 & 0.4351\tabularnewline
\hline 
METAMAP & 0.3 & 0.6838 & 0.4701 & 0.5441\tabularnewline
\hline 
\end{tabular}
\par\end{centering}
\caption{Performance evaluation of the proposed BERT+LDA model with METAMAP
and Amazon medical comprehend \label{tab:Performance-comparison-of-current-systems} }
\end{table}

We evaluate the performance of the proposed BERT+LDA model into two
different stages. In the first stage, we compare the performance of
the The BERT + LDA model the with state-of-the-art UMLS concept recognition
tool METAMAP \cite{aronson2006metamap} and clinical named entity
recognition model Amazon Comprehend Medical \cite{bhatia2019comprehend}.
In the second stage, we compare the BERT+LDA against Word2Vec+LDA. In both architectures we use Principal Coefficient Analysis (PCA) \cite{friedman2001elements} for dimensionality reduction of text vectors. We use Chain Classifier \cite{friedman2001elements} to frame the task as multi label classification.

\subsection{Experimental Setup}

We use curated clinical text to train and test our models.
All of the text instances contain at least the parent entity and
some contain a combination of time-onset-severity characteristics.
We use instances for the symptom ``headache'' in the following experiments.
Randomly sampled 80\% of the data is used for training and 20\% is
used for evaluation. 

The fine tuned BERT model has 24 layers and 1024 neurons per layer, activation function
used is Gaussian Error Linear Units(GELU) and the vocabulary size of 30522. LDA model is optimized
using a off-the-shelf Singular value decomposition (SVD) solver where
convergence limit is set to 1.0e-7. The Word2Vec \cite{mikolov2013distributed}
model is pre trained on Google News dataset and contains 3 million
words with each word having a vector of 300 dimensions.

\subsection{Performance Metric}

We use the following metrics:
\begin{enumerate}
\setlength\itemsep{0.4em}
    \item {$accuracy=\frac{correct\ predictions}{sample\ size}$}
    \item {$precession=\frac{TP}{TP +FP}$}
    \item {$recall=\frac{TP}{TP +FN}$}
    \item {$F_{1}score=2\times\frac{precession*recall}{precession+recall}$}
\end{enumerate}
Where TP is True Positive, FP is False Positive and FN is False Negative.

\subsection{Experimental Results }

\cref{tab:Performance-comparison-of-current-systems} shows
the comparison of the proposed BERT + LDA model with
the METAMAP and Amazon medical comprehend respectively. The BERT+LDA
model is superior than the Amazon medical comprehend about 3.8 times
on accuracy, 1.6 times on precession, 2.4 times on recall and 2 times
on $F_{1}$ score respectively. A similar superior result can be observed
against the METAMAP as well. 

\cref{tab:word2vec_bert} compares the performance of the BERT
+ LDA and Word2vec + LDA model respectively. The BERT + LDA is 32
times better on accuracy than the Word2vec + LDA model. In this experiment,
the precession, recall and $F_{1}$-score is improved for BERT+LDA
by a factor of 1.35, 3.32 and 2.4 times in comparison to the Word2vec
+ LDA respectively.

\begin{table}[t]
\begin{centering}
\begin{tabular}{|p{1.5cm}|p{1cm}|c|c|c|c|}
\hline 
Model & Contextual information & Accuracy & Precession & Recall & F1-score\tabularnewline
\hline 
WORD2VEC + LDA & NO & 0.0215 & 0.6842 & 0.2532 & 0.3696\tabularnewline
\hline 
BERT+ LDA & YES & \textbf{0.6428} & \textbf{0.9245} & \textbf{0.8376} & \textbf{0.8789}\tabularnewline
\hline 
\end{tabular}
\par\end{centering}
\caption{Performance comparison of word2vec and BERT\label{tab:word2vec_bert}}
\end{table}

In summary, the BERT + LDA performs superior in comparison to the
other entity recognition model and popular word embedding model the
Word2vec. This is because the BERT encodes the text capturing the context
of the entire text while Word2vec has a vector for each word in the text. When embedding the entire text using Word2Vec, word vector for each word in the text is averaged and in doing so the sequence of words is not taken into consideration.

To check the performance of the BERT + LDA model qualitatively, we plot the output of the 2 dimensional vectors generated by the LDA model.
\cref{fig:clusters}
shows how the related text is represented closer in vector space . For  the  Duration  (time),  the clusters of the BERT+LDA model are linearly separable, however, the clusters of the Word2Vec+LDA are overlapped. For the frequency (time) and the onset, the clusters in both models are clearly separable. For severity, the clusters in both models are partially overlapped, however, it seems that the configuration of the clusters in the BERT+LDA model is better than the Word2Vec+LDA model. Overall, the BERT+LDA model is superior to the Word2Vec+LDA model.

\section{Conclusion\label{sec:conclusion}}

Existing clinical named entity recognition
models are designed to predict the parent entity (eg. \emph{headache}) in a text input. However, these models fail to recognize the time-onset-severity
characterization of a parent entity (e.g. \emph{days} or \emph{months},
\emph{sudden} or \emph{gradual}, \emph{severe} or \emph{mild}). In this paper,
we have proposed a model which is a combination of a language understanding
framework and a classification method to predict both the
parent entity and the time-onset-severity characterization of the
parent entity. The proposed model successfully exploits the contextual
information of the parent entity to predict its time-onset-severity
characterizations. The proposed model has shown a superior performance
against the state-of-the-art  clinical named entity recognition frameworks METAMAP and Amazon
medical comprehend.

\bibliographystyle{IEEEtran}
\bibliography{example}

\end{document}